\documentclass[10pt, conference, compsocconf]{IEEEtran}

\usepackage{times}
\usepackage{epsfig}
\usepackage{graphicx}
\usepackage{amsmath}
\usepackage{amssymb}
\usepackage{graphicx}
\usepackage{graphics} 
\usepackage{float}
\usepackage{epsfig} 
\usepackage{mathptmx} 
\usepackage{times} 
\usepackage{amsmath} 
\usepackage{amssymb}  
\usepackage{booktabs}
\usepackage{epstopdf}
\usepackage{color}
\usepackage{url}
\usepackage{cite}

\usepackage[pagebackref=true,breaklinks=true,letterpaper=true,colorlinks,bookmarks=false]{hyperref}

\begin{document}
\title{Multi-Spectral Visual Odometry without Explicit Stereo Matching}

\author{Weichen~Dai$^{1}$,~Yu~Zhang$^{1}$,~Donglei~Sun$^{2}$,~Naira~Hovakimyan$^{2}$,~Ping~Li$^{1}$\\
$^{1}$Zhejiang University\\~$^{2}$University~of~Illinois~at~Urbana-Champaign \\
{\tt\small \{weichendai,zhangyu80\}@zju.edu.cn,~\{dsun13,nhovakim\}@illinois.edu,~pli@iipc.zju.edu.cn}
}

%
\maketitle

\begin{abstract}

Multi-spectral sensors consisting of a standard (visible-light)
camera and a long-wave infrared camera can simultaneously provide both visible and thermal images.
Since thermal images are independent from environmental illumination, they can help to overcome certain limitations of standard cameras under complicated illumination conditions. 
However, due to the difference in the information source of the two types of cameras,
their images usually share very low texture similarity. Hence, traditional texture-based feature matching methods cannot be directly applied to obtain stereo correspondences. To tackle this problem, a multi-spectral visual odometry method without explicit stereo matching is proposed in this paper.
Bundle adjustment of multi-view stereo is performed on the visible and the thermal images using direct image alignment.
Scale drift can be avoided by additional temporal observations of map points with the fixed-baseline stereo. Experimental results indicate that the proposed method can provide accurate visual odometry results with recovered metric scale. Moreover, the proposed method can also provide a metric 3D reconstruction in semi-dense density with multi-spectral information, which is not available from existing multi-spectral methods.

\end{abstract}

\section{Introduction}
\label{introduction}

Visual Odemetry (VO) methods have attracted significant attention since they can provide navigation information for autonomous missions in unknown environments. 
However, standard cameras that are commonly used in VO can only take advantage of the rich texture in good illumination conditions, and hence are inevitably influenced by strong illumination change.
For example, in firefighting and rescue scenarios, standard cameras cannot detect humans due to smog obscuration.
In such circumstances, Long-Wave Infrared (LWIR) cameras can not only provide the information through the smog but also capture thermal radiation without being affected by illumination. To combine the robustness of LWIR cameras and the rich-texture information from standard cameras, multi-spectral visual odometry methods have been developed on these two types of cameras \cite{mouats2015multispectral}. 

\begin{figure}[t]
	\centering
	\includegraphics[scale=0.28]{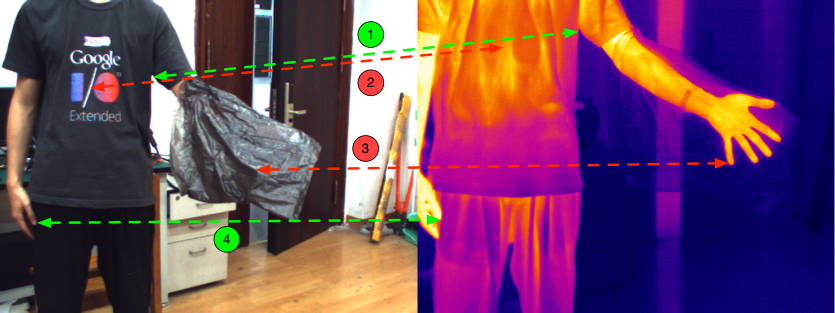}
	\caption{Potential (green) and unfeasible (red) correspondences.
	For potential correspondences~{\small\raisebox{.5pt}{\textcircled{\raisebox{-.9pt} {1}}}} and {\small\raisebox{.5pt}{\textcircled{\raisebox{-.9pt} {4}}}}, the gradient patterns between different correspondences are different.
	For unfeasible correspondences~{\small\raisebox{.5pt}{\textcircled{\raisebox{-.9pt} {2}}}} and {\small\raisebox{.5pt}{\textcircled{\raisebox{-.9pt} {3}}}}, there is no shared texture.
		Both types of stereo correspondences are difficult to find using stereo matching.
}
	\label{fig:diffintwospectrum}
\end{figure}

Due to the difference in information source, visible images and thermal images share little similar texture. This leads to the situation that two types of stereo correspondences, potential and unfeasible correspondences,  are difficult to deal with using explicit stereo matching. Figure~\ref{fig:diffintwospectrum} shows some examples of potential and unfeasible correspondences, depicted by the green and red dashed lines respectively. Existing explicit stereo matching can only extract potential correspondences associated with similar texture from the two images. As a result, even though each image contains rich texture, the VO methods based on stereo matching cannot yield robust estimation if only a few potential correspondences exist and the methods will even fail when there are no potential correspondences. The reliance on shared texture limits their performance and applicability in real-world environments. 

In this paper, a Multi-Spectral Visual Odometry (MSVO) method without explicit stereo matching is proposed.
It removes the aforementioned limitations of existing multi-spectral methods by recovering metric scale based on temporal stereo of cameras.
Meanwhile, the proposed method leverages the static stereo information by additional observations in the LWIR images.
Therefore, it can provide motion estimation results with recovered metric scale without using stereo correspondences.

The main contributions of this paper are as follows:
\begin{itemize}
	\item {A new multi-spectral visual odometry method without explicit stereo matching is proposed. Besides odometry results, the method can also provide a metric semi-dense 3D-reconstruction together with multi-spectral information for each map point, even if the two spectra share no similarity.}
	\item The new method overcomes the problem of temporary image unavailability of uncooled LWIR cameras.
	\item Quantitative and qualitative evaluations of the new method on a multi-spectral dataset are conducted. Experimental results indicate that without using explicit stereo matching the proposed method can still prevent metric scale drift and yield satisfactory odometry in challenging environments.
\end{itemize}

The rest of the paper is organized as follows. 
Related work is reviewed in Section \ref{sec:relatedwork}, and notations are defined in Section \ref{sec:notation}. 
The proposed system and procedures to retrieve the information of static stereo are discussed in Section \ref{sec:system}, and experimental results are presented in Section \ref{Experiment}.
Finally, conclusions are drawn in Section \ref{conclusion}.

\section{Related work}

\label{sec:relatedwork}

Visual odometry methods based on standard cameras were first proposed by Nister \cite{nister2004visual} and can be categorized into filter-based \cite{chiuso2002structure,davison2007monoslam} and keyframe-based methods \cite{klein2007parallel, mouragnon2006real}. The latter can be divided into two categories: feature-based methods \cite{forster2017svo,mur2017orb} and direct methods \cite{newcombe2011dtam,engel2014lsd,engel2017direct}. 
The distinction between those two methods is that feature-based methods use reprojection error of feature points, while direct methods exploit photometric error of raw images directly.

Nowadays, researchers have tried to extend existing VO methods to exploit infrared sensors \cite{lin2001review}.
Among infrared sensors, LWIR spectrum has the largest difference compared with the visible spectrum, and considering their price advantage, uncooled LWIR cameras show great potential to enhance the robustness of visual odometry. Though economically accessible, uncooled cameras have their own shortcomings as  mentioned in \cite{hajebi2008structure},
among which  the following three are worth to mention.
(1) High Noise: there is fixed noise in the image due to camera self-emission. For this reason, most uncooled cameras use Non-Uniformity Correction (NUC) to eliminate the fixed noise.
(2) Special Mechanism (NUC corruption): for every ten seconds LWIR cameras need about half second to do NUC, during which the images are corrupted and hence unavailable.
(3) High Dynamic Range: since the imaging strength of an object is approximately proportional to the fourth order of its temperature, objects with large temperature difference will lead to image saturation and hence high frequency information loss.
Due to these factors, visual odometry with monocular \cite{jung2007egomotion, borges2016practical, mouats2015thermal} and stereo \cite{owens1999passive, rankin2011unmanned} LWIR setups cannot provide results as accurate as visual odometry based on visible information in most environments.
Hence, multi-spectral visual odometry methods using LWIR as a complementary information source have attracted much attention \cite{mouats2015multispectral}, since this setup has the potential to work in poorly illuminated environments.

\begin{figure}[t]
	\centering
	\includegraphics[scale=0.25]{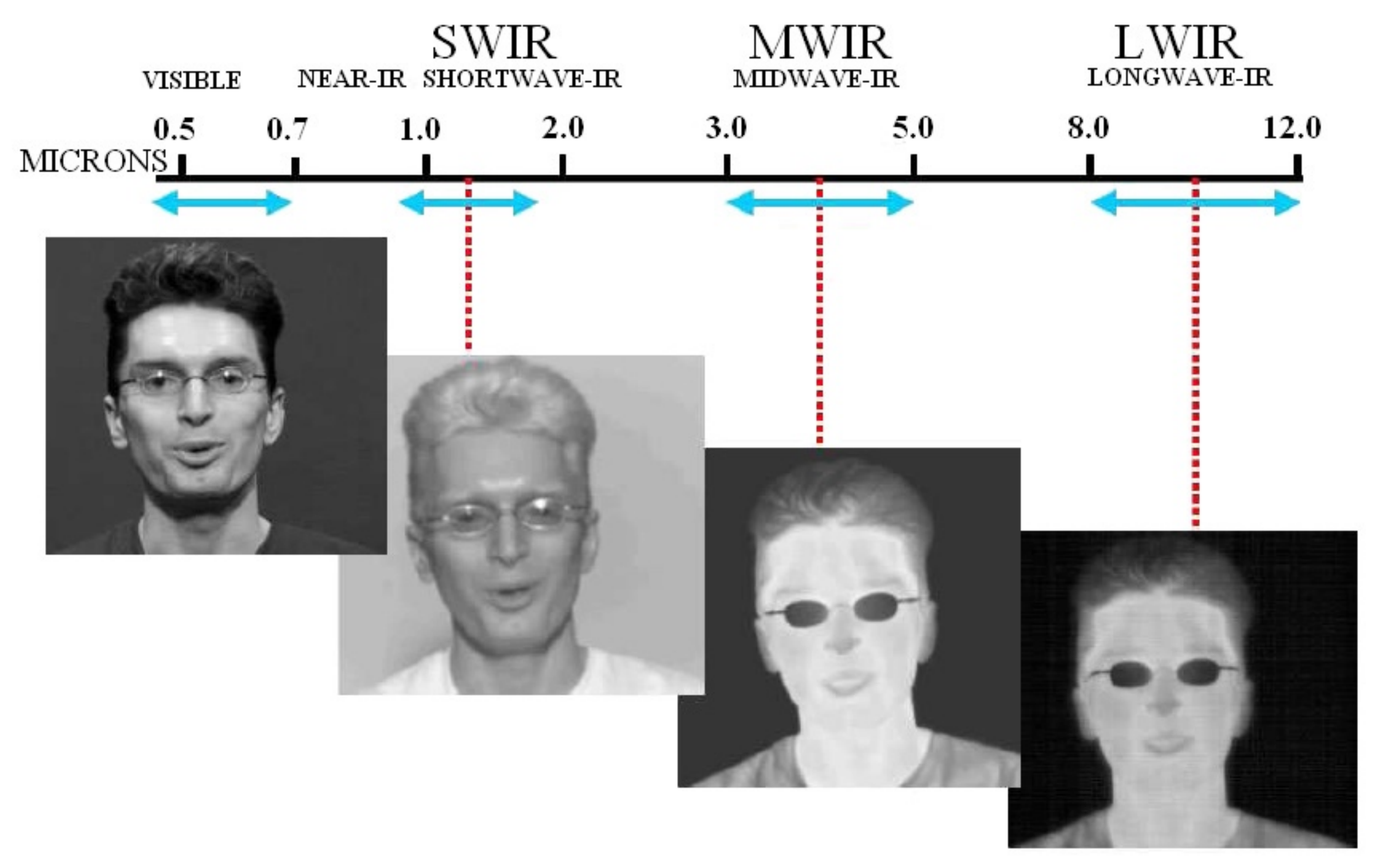}
	\caption{
		The IR images of a human head in different wavelengths \cite{hajebi2008structure}. The difference between the far infrared (LWIR) image and the near infrared (SWIR) image shows the independence of reflectance in the former. Furthermore, the LWIR and MWIR lights cannot pass through glasses \cite{arnell2005vision}.
	}
	\label{fig:diffwave}
\end{figure}

Multi-spectral visual odometry using both standard and LWIR cameras can be seen as a special new type of stereo setup. 
Due to different modalities shown in Figure~\ref{fig:diffwave}, classic stereo matching cannot be applied directly to obtain stereo correspondences. For this reason, existing methods try to tackle the problem from two angles: monocular visual odometry and stereo visual odometry.
The first type of methods treat the multi-spectral setups as a monocular sensor by ignoring the parallax and superimpose the LWIR image directly onto the visible image to obtain stereo correspondences \cite{magnabosco2013cross,poujol2016visible,chen2017rgb}. However, not only do these methods introduce errors due to the parallax that shouldn't have been ignored, they also give up the information of static stereo, leading to scale drift in estimation results.
The other type of methods leverage special matching algorithms
to obtain as many potential stereo correspondences as possible, with which modern binocular feature-based methods can be exploited
\cite{mouats2015multispectral,beauvisage2016multi,pinggera122012cross}. However, as described in Section~\ref{introduction}, this type of methods can only obtain part of potential stereo correspondences.
Meanwhile, the explicit stereo matching used in these methods abandons unfeasible correspondences, though they might still contain multi-view stereo relations.
Hence, research on multi-spectral visual odometry is still in the preliminary stage.

The observations in binocular cameras with fixed baseline should have multi-view stereo relations. Therefore, the photometric bundle adjustment \cite{engel2017direct,engel2014lsd,newcombe2011dtam} of multi-view stereo is used to achieve multi-spectral visual odometry.
In the proposed method, static stereo information is obtained by performing calculations of photoconsistency only between temporal stereo of each camera  but not between the two cameras of the stereo setup.
The reason for using photoconsistency is that the proposed method needs to exploit all types of gradient because map points may not be projected on the salient features such as corners or edges in the other camera.
A similar idea of individual calculation has also been used in other methods for solving the non-overlapping field of view (FOV) problem \cite{kim2010motion,clipp2008robust,kazik2012real}. 
The proposed method not only obtains stereo constraint by cross-projection of map points, but also employs the motion consistency used by methods based on non-overlapping FOV. 
Therefore, the proposed method can obtain more accurate results since more constraints are considered.

\section{Notations}

\label{sec:notation}

Before the proposed MSVO algorithm is discussed in detail, some notations and symbols will be defined first.
A standard and a LWIR cameras, which are fixed on a tripod and can take images simultaneously, are indexed by $a$ and $b$ and denoted by $c_a$ and $c_b$ respectively.
The intensity images collected at the sampling time $k$ by those two cameras are denoted by $\mathbf{I}_k^a$ and $\mathbf{I}_k^b$, where $\mathbf{I}_k^a,\mathbf{I}_k^b:\mathbb{R}^2 \mapsto \mathbb{R}$.
The $i$-th 3D point $\mathbf{p}_{i}^w$ in the world frame $w$ is projected to $\mathbf{u}_{i}^{a}$ in the image coordinates of $c_a$ with a transformation $\mathbf{T}_{k,w} \in SE(3)$ using
\begin{equation}
\mathbf{u}_{i}^{a} = \pi_{a}(\mathbf{T}_{k,w} \cdot \mathbf{p}_{i}^w),
\end{equation}
where $\pi_{a} \colon \mathbb{R}^3 \mapsto \mathbb{R}^2$ is the camera model of $c_a$, and the superscript $a$ indicates that $\mathbf{u}_{i}^{a}$ is in the image coordinates of $c_a$, assuming that the optical center of $c_a$ is set as the origin of the body frame.

For the transformation $\mathbf{T}_{k,w}$, the Li Algebra $\mathfrak{se}(3)$ corresponding to the tangent space of $SE(3)$ is used in the optimization. 
A vector $\zeta_{k,w} \in \mathbb{R}^6$ can be mapped to $SE(3)$ by the exponential mapping through
\begin{equation}
\exp(\zeta_{k,w}^\land)=\mathbf{T}_{k,w},
\end{equation}
where the $^\land$ operator transforms elements in $\mathbb{R}^6$ into elements in $\mathbb{R}^{4\times4}$.
The transformation between the~$k$-th and the $l$-th frame is denoted by the equation $\mathbf{T}_{l,k} = \mathbf{T}_{l,w} \mathbf{T}_{k,w}^{-1}$. 
The result of the map point reprojected to the target image frame is denoted by a superscript $'$, such as $\mathbf{u}_{i}'^{a}$.
In addition, the motion between the reference and the target frame is denoted by $\mathbf{T}'$.

\section{Multi-spectral visual odometry}
\label{sec:system}

\label{sysoverview}

\begin{figure}[t]
	\centering
	\includegraphics[scale=0.15]{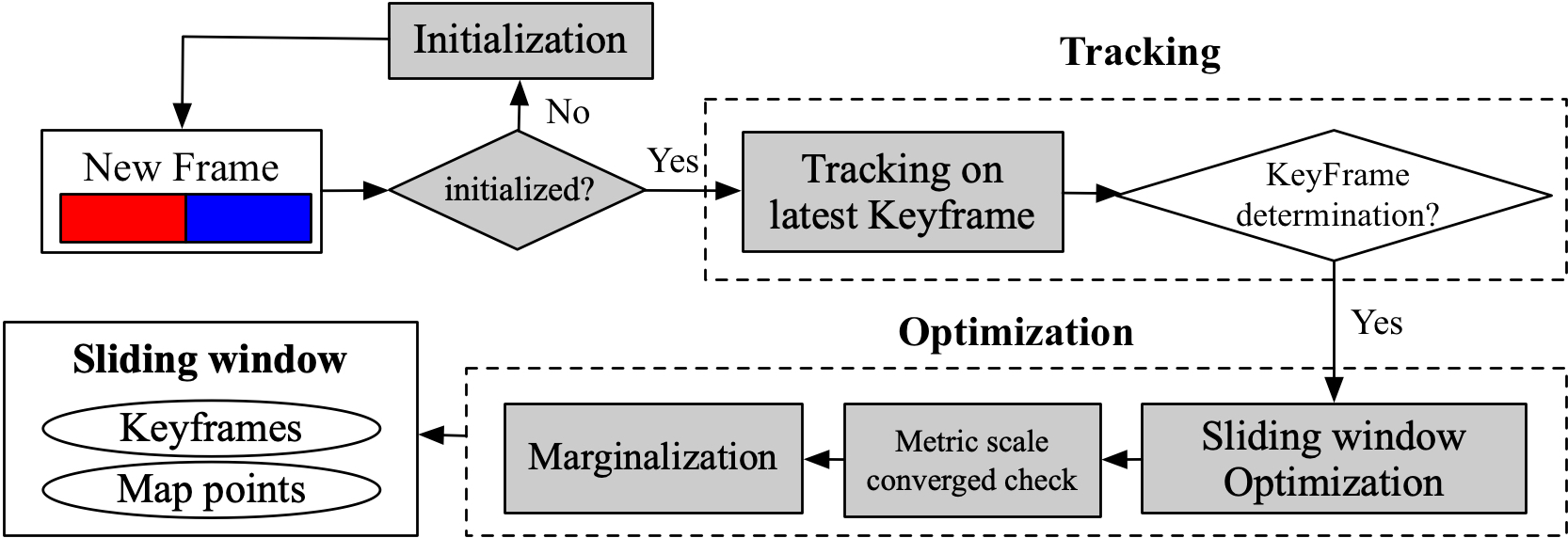}
	\caption{MSVO system overview. The red and blue color denote two types of images from different modalities.
		The steps in gray boxes are modified based on classic direct methods\cite{engel2017direct}.}
	\label{fig:overview}
\end{figure}

The structure of the proposed multi-spectral visual odometry is depicted in Figure~\ref{fig:overview} and includes three components: initialization, tracking, and optimization.

Different from classic stereo methods \cite{wang2017stereo}, although the setup is itself a binocular, the map points have to be initialized based on random depth, since stereo matching is not used in the proposed method. Consequently, the initialization can only provide a result up to a scale and the metric scale has to be recovered in the following processes. 

If the initialization succeeds, each new frame containing multi-spectral data is first tracked to the latest keyframe in the coarse-to-fine manner based on the direct image alignment formulation.
If the scene observed from the new frame has a large difference from that observed from the keyframe, this new frame will be added to the sliding window as a new keyframe. 

In the sliding window, all measurements are used to optimize keyframe poses, affine brightness parameters, and the inverse depth of observed map points, again based on the direct image alignment of multi-view stereo.
Only the photoconsistency between the temporal views of each camera is calculated, and the photoconsistency between the two cameras of the stereo setup is not needed.
After the sliding window optimization, a convergence check will be conducted to verify whether the metric scale has converged.
To maintain the size of the sliding window, old keyframes and map points will be marginalized using Schur complement.

\subsection{Initialization}
\label{sec:init}

\begin{figure}[tpb]
	\centering
	\includegraphics[scale=0.18]{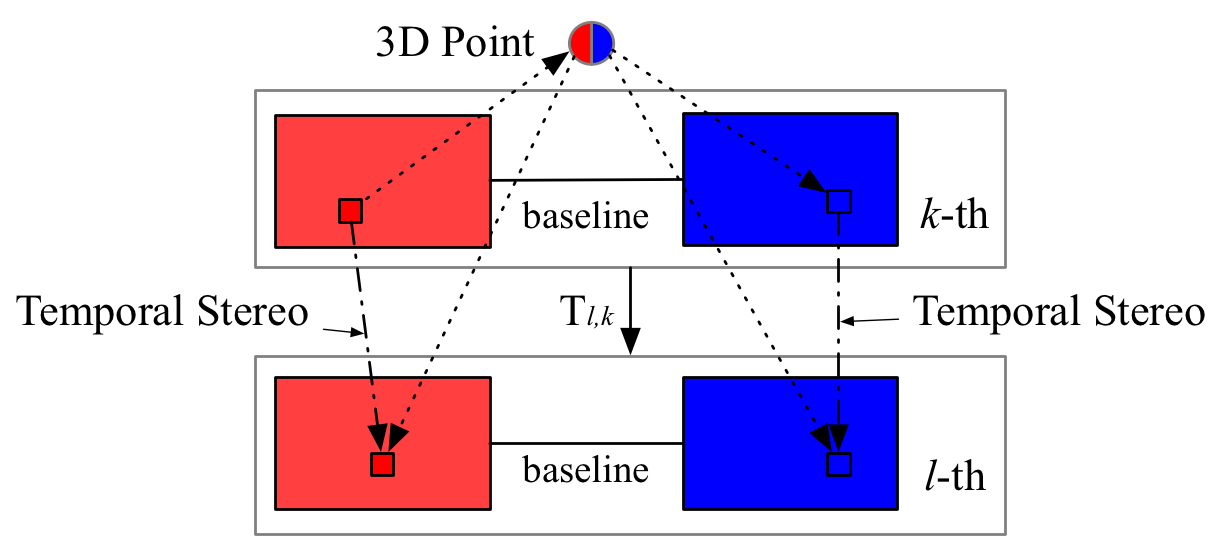}
	\caption{One point with two temporal matching results. The red and the blue colored data are from cameras of different types. The photometric residual is only calculated between the two pixel coordinates from the same camera source.}
	\label{fig:relation}
\end{figure}

As shown in Figure~\ref{fig:relation}, the proposed method does not use explicit stereo matching. To obtain static stereo information directly through the two images of one frame is not feasible. 
Therefore, methods based on random depth are used in the initialization step \cite{engel2013semi,engel2014lsd,engel2017direct}.
Then, the created map points are scaled such that the mean inverse depth is one.
The metric scale cannot be directly obtained in initialization either, and it will be recovered during the sliding window optimization. The process of recovering the metric scale can be seen as an additional step of initialization.

\subsection{Direct image alignment formulation}
The direct image alignment formulation of a map point in a reference frame ($\mathbf{I}_{k}^{a}$, $\mathbf{I}_{k}^{b}$) observed in a target frame ($\mathbf{I}_{l}^{a}$, $\mathbf{I}_{l}^{b}$) can be divided into two types: 
the formulation of the temporal multi-view stereo from the point reprojected into the subsequent frames of the standard camera, and the formulation of the static stereo with additional temporal observations from the point reprojected into those two frames of the LWIR camera. These two formulations will be discussed in the following.

\subsubsection{Temporal multi-view stereo}

For the $i$-th map point, the direct image alignment of temporal multi-view stereo can be formulated as 
\begin{equation}
\label{equ:classic_cost}
\mathbf{E}_{l,k}^{i,a} = 
\left\Vert
(\mathbf{I}_l^{a}(\mathbf{u}_{i}'^{a})- \beta_{l,a})
-
\frac{e^{\alpha_{l,a}}}{e^{\alpha_{k,a}}}
(\mathbf{I}_{k}^{a} (\mathbf{u}_{i}^{a}) - \beta_{k,a})
\right\Vert_\delta,
\end{equation}
where $\alpha_{k,a}$, $\alpha_{l,a}$, $\beta_{k,a}$, and $\beta_{l,a}$ are parameters of the affine brightness transfer function which improves robustness of the system in environments with dynamic illumination, and $||.||_\delta$ is the Huber norm, while
$\mathbf{u}_{i}'^{a}$ is obtained using
\begin{equation}
\mathbf{u}_{i}'^{a} = \pi_{a} (\mathbf{T}_{l,k} \mathbf{p}_{i}^{a}),
\end{equation}
where
$\mathbf{p}_{i}^{a}$ is defined as an indicating vector parameterized by only the inverse depth $d_{i}$ as
\begin{equation}
\mathbf{p}_{i}^{a} = \pi_{a}^{-1}(\mathbf{u}_{i}^{a},d_{i}).
\end{equation}
This is the classic direct image alignment with a fixed reference intensity $\mathbf{I}_{k}^{a} (\mathbf{u}_{i}^{a})$ that is used in direct methods. 
The estimation result obtained from only direct image alignment of temporal multi-view stereo is lacking in metric scale information. Therefore, it is necessary to use the information of static stereo to prevent the scale from drifting, which is introduced next.

\subsubsection{Static stereo}

In order to obtain stereo information from the static stereo, the direct image alignment formulation of the static stereo $\mathbf{E}_{l,k}^{i,b}$ between the two frames of the LWIR camera is defined as

\begin{equation}
\label{equ:static_stereo_cost}
\mathbf{E}_{l,k}^{i,b}= 
\left\Vert
(\mathbf{I}_l^{b}
(\mathbf{u}_{i}'^{b}) - \beta_{l,b})
-
\frac{{e^{\alpha_{l,b}}}}{e^{\alpha_{k,b}}}
(\mathbf{I}_{k}^{b}
(
\mathbf{u}_{i}^{b}
) - \beta_{k,b})
\right\Vert_\delta,
\end{equation}
where $\mathbf{u}_{i}^{b}$ is given by
\begin{equation}
\mathbf{u}_{i}^{b} = \pi_{b}( \mathbf{p}_{i}^{b} )= \pi_{b} (\mathbf{T}_{b,a} \mathbf{p}_{i}^{a}),
\end{equation}
and $\mathbf{u}_{i}'^{b}$ can be computed from
\begin{equation}
\mathbf{u}_{i}'^{b} = \pi_{b} (\mathbf{p}_{i}'^{b}) = \pi_{b} (\mathbf{T}_{b,a} \mathbf{T}_{k,l} \ \mathbf{p}_{i}^{a}).
\end{equation}

Note that in this direct image alignment of static stereo, both the value of reference intensity and of target intensity need to be obtained from the projection of $\mathbf{p}_{i}^{a}$ in the other camera. This individual observation of temporal stereo relaxes the requirement of classic binocular visual odometry methods on stereo matching.

\subsubsection{Observability}
\label{sec:obs}

\begin{figure}[t]
	\centering
	\includegraphics[scale=0.08]{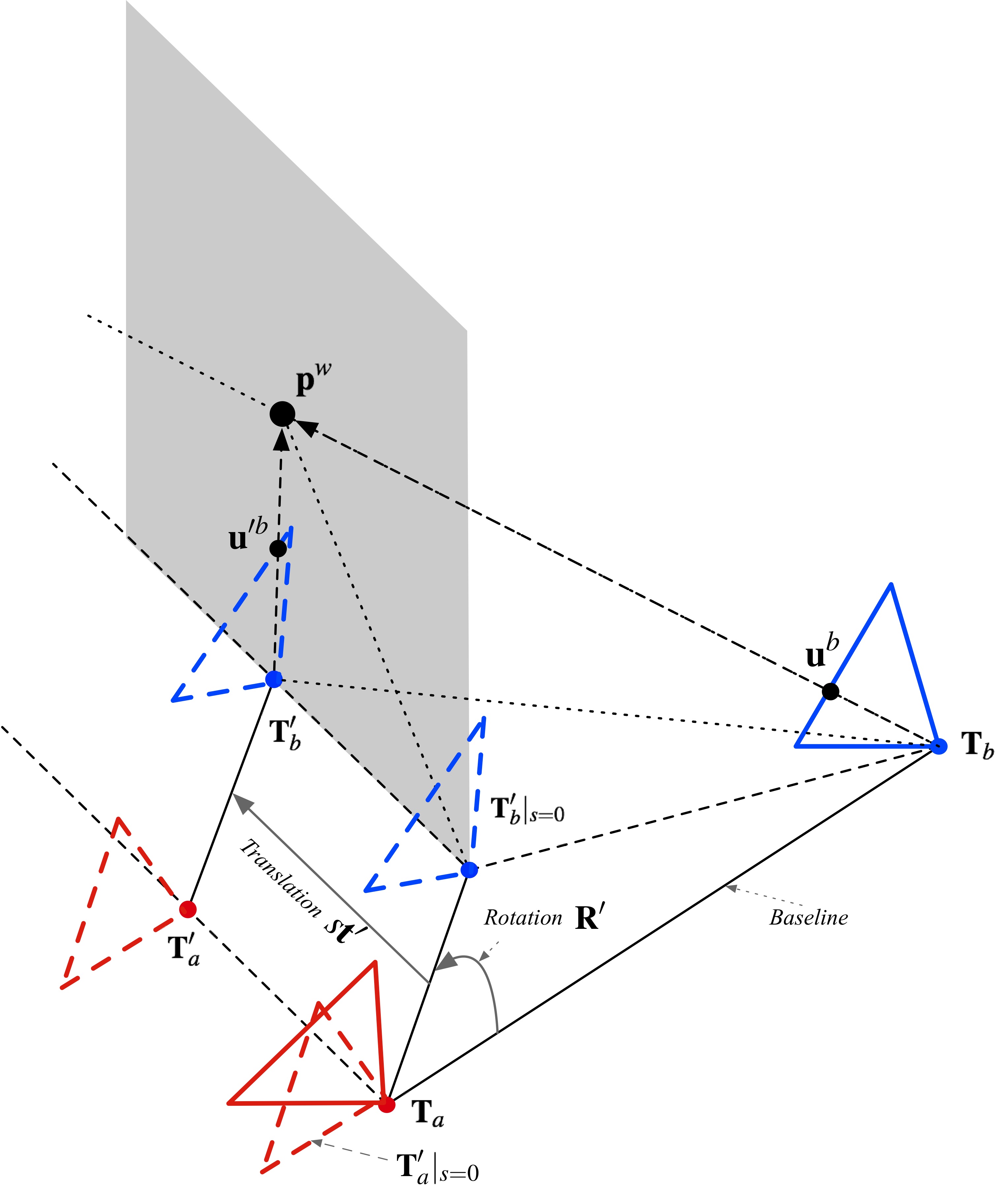}
	\caption{The observability of metric scale. The red and the blue color denote two cameras. The solid triangles indicate camera poses in the reference frame while the dashed triangles indicate the camera poses after a transformation. On the gray plane, the position of point $\mathbf{p}^w$ must be at the intersection of those two rays from the center of two frames. }
	\label{fig:geo_interp}
\end{figure}

In order to show the observability of the metric scale using the proposed direct image alignment, we present an example with two frames and one map point.
As shown in Figure~\ref{fig:geo_interp}, the poses of the cameras at the starting and the subsequent sampling time-stamps are denoted by $\mathbf{T}_a$, $\mathbf{T}_b$ and $\mathbf{T}'_a$, $\mathbf{T}'_b$ respectively.
Assume that  $\mathbf{T}_a$ is at the origin of the world frame $w$.
Then based on the baseline transformation $\mathbf{T}$, we have \[\mathbf{T}_a = \left[ \begin{matrix} \mathbf{I} & \mathbf{0} \\ \mathbf{0}^\top & 1 \end{matrix} \right]\quad\text {and}\quad \mathbf{T}_b = \left[ \begin{matrix} \mathbf{R^\top} & -\mathbf{R^\top}\mathbf{t} \\ \mathbf{0}^\top & 1 \end{matrix} \right].\] Let 
$\mathbf{t}_b$, $\mathbf{t}'_b$, and $\mathbf{t}'^*_b $  be the position vector of $\mathbf{T}_b$,  $\mathbf{T}'_b$, and $\mathbf{T}'_b|_{s=0}$ respectively which satisfy the following equations: $\mathbf{t}'_b = \mathbf{t}+s\mathbf{R}\mathbf{t'}$, $\mathbf{t}'^*_b = -\mathbf{R}'^{-1}\mathbf{R}^{-1}\mathbf{t}$, and $\mathbf{t}_b = -\mathbf{R}\mathbf{t}$, where $s$ is the unknown metric scale.

The motion between two time-stamps that leads to an unobservable metric scale is called a critical motion. In the following, two critical conditions will be discussed.
The first critical condition is satisfied when $\mathbf{t}'_b$ lies on the line passing through $\mathbf{t}_b$ and $\mathbf{t}'^*_b $, i.e. when
\begin{equation}
\label{eq:non-obs2}
\begin{aligned}
\mathbf{t}'^\top_b(\mathbf{t}_b\times\mathbf{t}'^*_b ) = 0,
\end{aligned}
\end{equation}
which implies that $\mathbf{t}'_b$, $\mathbf{t}_b$, and $\mathbf{t}'^*_b $ are unable to form a unique triangle.
The other critical condition comes from the map points. The point $\mathbf{p}^w$ is the intersection of the two rays passing through $\mathbf{u}^b$ and $\mathbf{u}'^b$ from the camera center of two consecutive frames in the world coordinates.
A unique tetrahedron is formed by $\mathbf{t}_b$, $\mathbf{t}'_b$, $\mathbf{t}'^*_b $, and  $\mathbf{p}^w$ if they are not on the same plane as shown in Figure~\ref{fig:geo_interp}.
When $\mathbf{t}_b$ is on the surface formed by $\mathbf{t}_b$, $\mathbf{t}'_b$, and $\mathbf{p}^w$, the metric scale parameter $s$ can be selected arbitrarily since an infinite number of similar triangles will be created.
As a result, the critical condition becomes
\begin{equation}
\label{eq:non-obs1}
\begin{aligned}
\left[
\begin{matrix}
(\mathbf{p}^w-\mathbf{t}'_b) \times (\mathbf{t}'^*_b-\mathbf{t}'_b) \\
-\mathbf{t}'^\top_b(\mathbf{t}_b\times\mathbf{t}'^*_b)
\end{matrix}
\right]^\top
\begin{bmatrix}
\mathbf{t}_b \\ 1
\end{bmatrix}
= 0
\end{aligned}.
\end{equation}
If Equation~\eqref{eq:non-obs1} holds, the metric scale $s$ is unobservable.

In order to avoid the situation of metric scale being unobservable, 
the recovery of the metric scale will be conducted in the sliding window.
Since there are enough points and multiple keyframes in the sliding window, the critical condition~(\ref{eq:non-obs1}) usually won't hold and the metric scale will be observable.

\subsection{Tracking}

As mentioned in Sections \ref{sec:init} and \ref{sec:obs}, the scale won't have been recovered in the system after the initialization, and the direct image alignment of static stereo has critical conditions.
Consequently, before the metric scale is fully recovered, in the tracking process all map points are projected into new frames. The cost function can be formulated as  
\begin{equation}
    \mathbf{E}_{l,k}^{a} = \sum_{i}\mathbf{E}_{l,k}^{i,a}.
\end{equation}
This equation is minimized while fixing the depth of all the map points to estimate the initial pose of the new frame.

The reason why Equation \eqref{equ:static_stereo_cost} is not used is that it will affect estimation accuracy of the tracking process with an unrecovered scale. Once the scale is fully recovered, the two direct image alignment formulations \eqref{equ:classic_cost} and \eqref{equ:static_stereo_cost} will be combined as a new cost function
\begin{equation}
    \mathbf{E}_{l,k}= \sum_{i}(\mathbf{E}_{l,k}^{i,a}+\mathbf{E}_{l,k}^{i,b})
\end{equation}
to be used in the optimization for motion estimation.

\subsection{Optimization}
\label{opt}

\begin{figure}[tpb]
	\centering
	\includegraphics[scale=0.08]{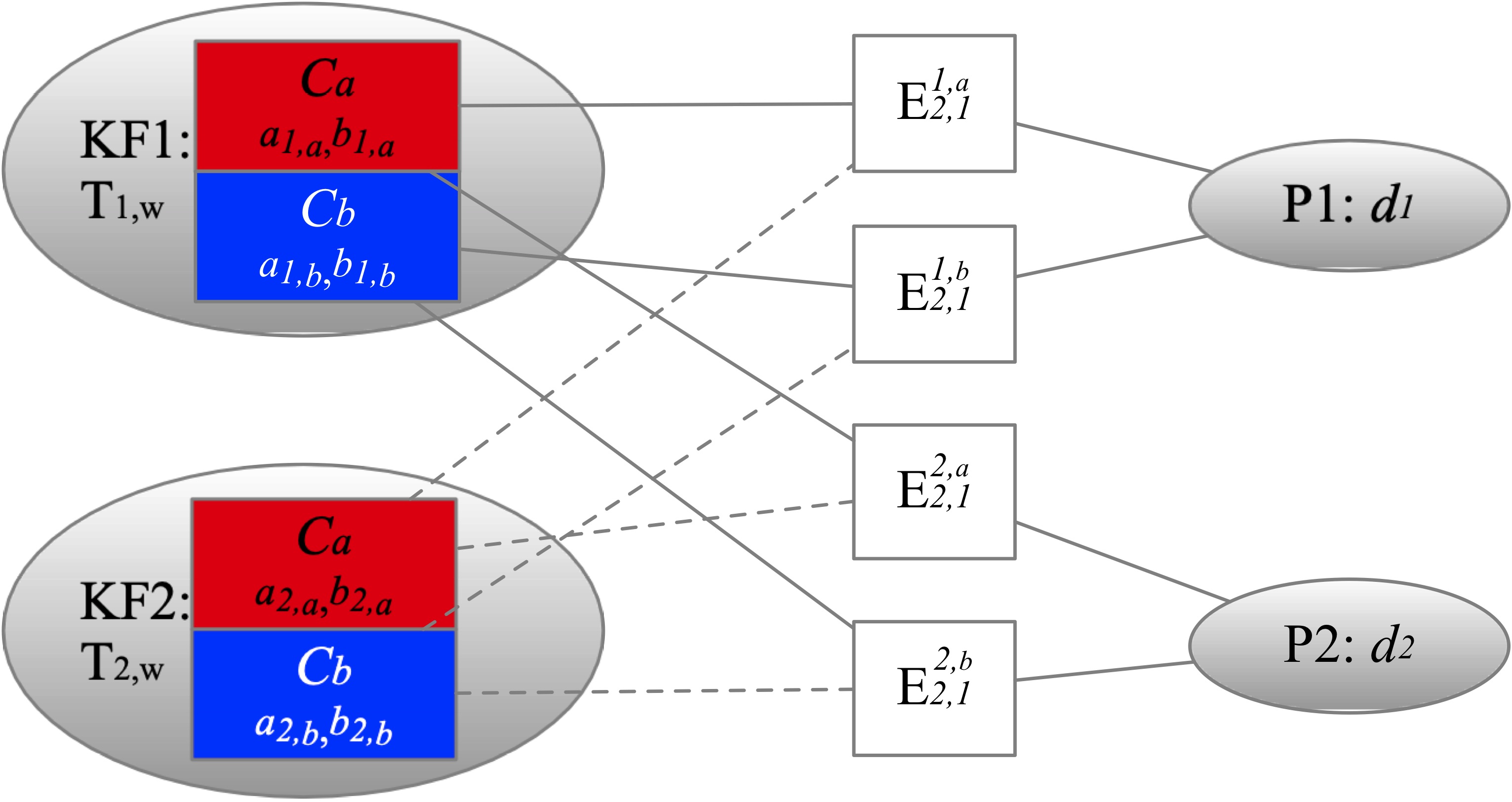}
	\caption{Factor graph for the full cost function. The red and the blue colored data are from the cameras of different types. Two map points are initialized from different cameras and observed in two keyframes. Each factor is related to one point and two keyframes of the same camera. Constraints from the initialized keyframes and the observed keyframes are represented in solid and dashed lines respectively.}
	\label{fig:factorgraph}
\end{figure}

As shown in Figure~\ref{fig:relation}, the same 3D point is initialized from one frame and projected to four pixel coordinates in four images where the photometric residual is calculated between the two frames from the same camera.
Thus, the full photometric error function over all the frames and points can be
found by
\begin{equation}
\label{equ:all}
\mathbf{E}_{full} = \sum_{k \in \mathcal{F}}\sum_{l \in (\mathcal{F} \char`\\ k)}  \mathbf{E}_{l,k},
\end{equation}
where $\mathcal{F}$ is the set of active keyframes in the sliding window.
The function  $\mathbf{E}_{full}$ in Equation (\ref{equ:all}) is briefly depicted in Figure~\ref{fig:factorgraph} where two keyframes and two map points are considered as an example. 
This error function is used to optimize all variables in the sliding window with the Gauss-Newton algorithm
\begin{equation}
\mathbf{\delta} = -\mathbf{H}^{-1} \mathbf{b},
\end{equation}
where $\mathbf{H} = \mathbf{J}^T \mathbf{W}\mathbf{J}$ and $\mathbf{b} = \mathbf{J}^T \mathbf{W}r$, with $\mathbf{W}$ being the diagonal weight matrix, $r$ the residual, and  $\mathbf{J}$ the Jacobian of all state variables computed from perturbations.
The new current state is updated incrementally by
\begin{equation}
\mathbf{f} \leftarrow \mathbf{\delta} \boxplus \mathbf{f},
\end{equation}
where $\mathbf{f} \in \mathfrak{se}(3)^n \times \mathbb{R}^m $ denotes the variables to be estimated through optimization, including all geometric variables $\{\mathbf{T}_{k,w}$, $\mathbf{T}_{l,w}$, $d\}$ and all photometric variables $\{(\alpha_{k,a}$, $\beta_{k,a}$, $\alpha_{l,a}$, $\beta_{l,a})$,  $(\alpha_{k,b}$, $\beta_{k,b}$, $\alpha_{l,b}$, $\beta_{l,b})\}$. The $\boxplus$ operator is defined as the operation \[\mathbf{\delta} \boxplus \mathbf{f} = exp(\mathbf{\delta}^\land) \cdot \mathbf{T}_{(c,w)}\] on $\mathfrak{se}(3) \times SE(3) \rightarrow SE(3)$, and the remaining parameters are updated by addition.

A convergence test is performed to check whether the metric scale has converged.
Geometric variables before and after the optimization are used to calculate scale change. 
If the scale change is less than a threshold, the scale recovery is considered complete.

After the sliding window optimization, marginalization is performed to remove old variables and maintain computational efficiency using Schur complement~\cite{leutenegger2015keyframe, engel2017direct}. To marginalize a keyframe and keep the special sparsity of $\mathbf{H}$, all map points in the keyframe and the points not observed in the latest two keyframes of the camera where these points were created are marginalized first. Then the keyframe is marginalized and removed from the sliding window.
Note that similar to the tracking process, Equation~\eqref{equ:static_stereo_cost} is not used until the metric scale is recovered.

\subsection{The NUC problem}
The NUC corruption problem can be solved with the proposed method as follows. Duing NUC, since the temporal residual of the LWIR camera is large, the corresponding frames will be considered corrupted and ignored.
Hence, when the LWIR camera enters NUC process, the tracking process will only use the latest data from the standard camera to estimate the latest egomotion.

\section{Experiments}
\label{Experiment}

This section aims to verify that the proposed algorithm is capable of fusing multi-spectral information to obtain motion estimation results, the metric scale, and stereo data associations.
Further, it indicates that the method yields a feasible visual odometry solution using the non-visible light spectrum while overcoming the defects of LWIR cameras mentioned earlier in Section~\ref{sec:relatedwork}.

In a multi-spectral dataset, the proposed method is evaluated in both indoor and outdoor environments to demonstrate that it can work with different sensors, recover the metric scale, and obtain a fused semi-dense map with correct stereo correspondences.
Since there are no benchmark or open-source algorithms applicable to multi-spectral configuration, a multi-spectral device and a dataset have to be built for the purpose of evaluation. 

\subsection{Multi-spectral device}
\begin{figure}[t]
	\centering
	\includegraphics[scale=0.035]{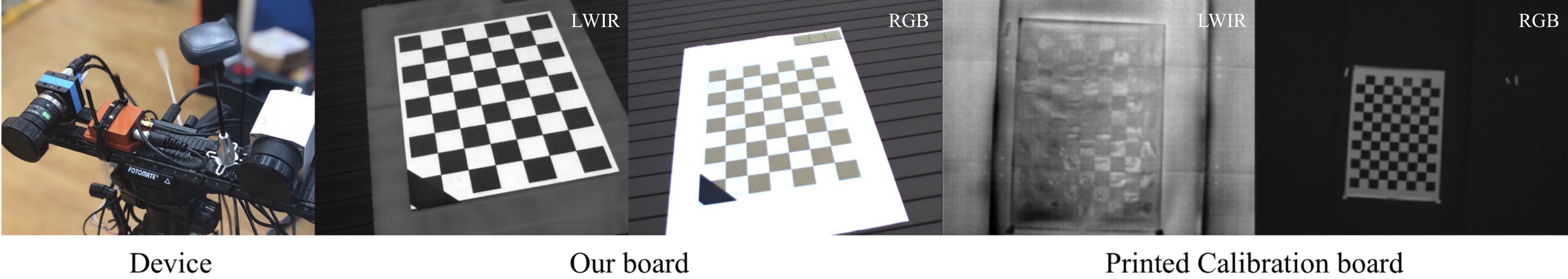}
	\caption{Multi-spectral device: The sensors from left to right are RGB camera, motion capture device, antenna, and LWIR camera. The second and third images are our calibration board and the rightmost two images are the printed board.}
	\label{fig:calib_sample}
\end{figure}

As shown in Figure~\ref{fig:calib_sample}, a system with a GPS receiver, a standard camera, and a LWIR camera has been developed.
The two cameras are hardware synchronized.
The result from GPS is considered as comparison data for outdoor environments.
In addition, instead of using a printed chessboard, a special chessboard whose surface is made of different materials is used to calibrate the system.
In the image sequences, the thermal image is captured on the LWIR camera with a rolling shutter, a small focus range, and a long exposure time. 
As a result, in addition to NUC corruption, the image also contains significant geometric noises (e.g. unfocused blur, geometric distortions, and motion blur).
Hence, the dataset recorded by this handheld device is very challenging for egomotion estimation.

\label{sec:calib}

\subsection{Quantitative evaluation}

\begin{figure}[t]
	\centering
	
	\includegraphics[width=0.45\textwidth]  {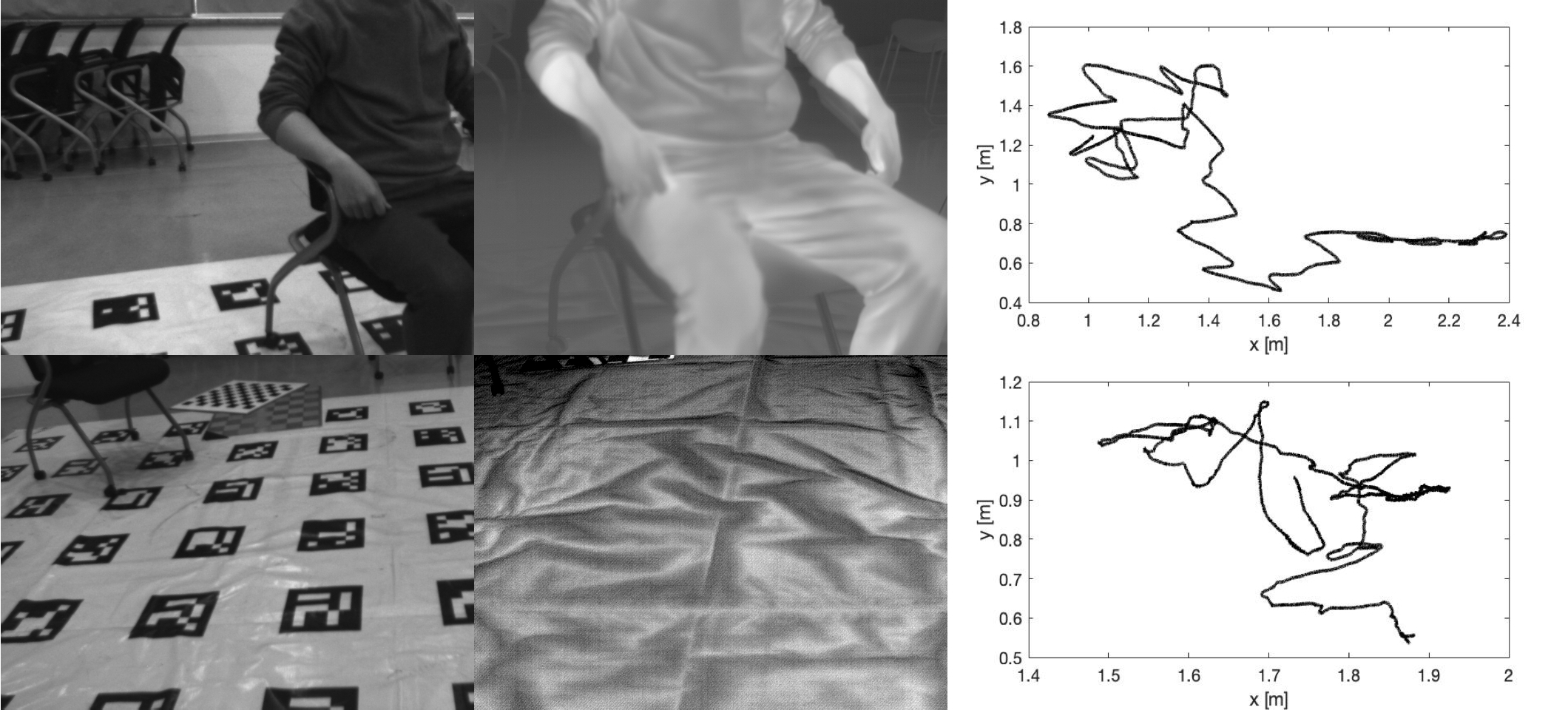}

	\caption{Two examples of the sequences in the dataset. The left column shows the RGB images, the middle column shows the thermal images, and the right column shows the ground truth trajectories. One sequence shows the environment with a person, and the other without.}
	\label{pic:envs}
\end{figure}

In order to make a quantitative evaluation, the proposed method was evaluated on a multi-spectral dataset with groundtruth from the motion capture system.
The dataset, as summarized in Table~\ref{tab:intro_dataset}, consists of 7 sequences that were recorded in different environments as shown in Figure~\ref{pic:envs} with corresponding camera trajectories. In the beginning part of each sequence, the device was translated and rotated simultaneously to avoid the critical motion and facilitate metric scale recovery.
Each object was put far away from the camera since both cameras have limited angle of view.
Considering that the thermal variance was small in the environment, several sequences containing a sitting person were also recorded to enrich the texture.
\begin{table}[t]
	
	\centering
	\caption{List of multi-spectral sequences
	}
	\label{tab:intro_dataset}
	\resizebox{8cm}{!}{
	\begin{tabular}{c|ccc}
			\toprule
			Method &  Duration [s]  & Avg. Trans. Vel. [m/s]  &  Avg. Rot. Vel. [deg/s]   \\
			\midrule
			sq-01   &{113.81}   & 0.0688      & 29.1738  \\
			sq-02   &{89.19} 	& 0.0690         & 39.6421  \\
			sq-03   &{88.99} 	& {0.0581}     & 33.0630  \\
			sq-04  &{102.44}   & {0.0969}     & 39.5428  \\
			sq-05   & 110.47	  & 0.0724        &  {28.6348} \\
			sq-06  &{159.19}   & {0.0575}         & 45.4522  \\
			sq-07  &{60.94}   & {0.1162}         & 31.6708  \\
			\bottomrule
	\end{tabular}}
	
\end{table}

\begin{figure}[t]
	\centering
	\includegraphics[scale=0.3]{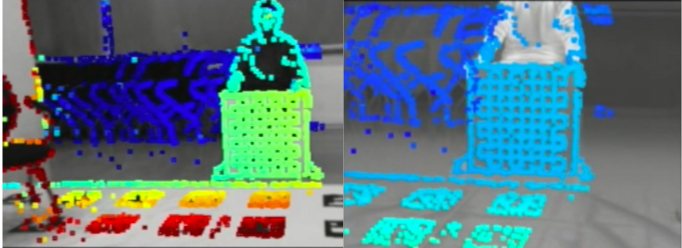}
	\caption{Depth map of multi-spectral images on Sq7.
	The right image shows the LWIR depth map.
		Each colored point in the depth map is obtained from the reprojection of that map point.
		The color indicates the distance from the 3D point to the optic center of that camera.
		The warmer the color, the smaller the distance.
	}
	\label{fig:data_ass_vicon}
\end{figure}

The proposed method was evaluated against two state-of-the-art methods: ORB-SLAM2 \cite{mur2017orb} and DSO \cite{engel2017direct}.
It should be noted that loop closing was turned off in ORB-SLAM2, while global bundle adjustment was still used. 
To compare the performance of different approaches, the results of ORB-SLAM2 and DSO are aligned with the groundtruth trajectories. 
As shown in Table \ref{tab_msd}, the Euclidean distance is computed between the estimation results and the groundtruth with translational Root Mean Square Error (RMSE) of Absolute Trajectory Error (ATE) in meters.
As shown in the RGB results, the metric scale of ORB-SLAM2 drifts away for most of the sequences since there are wrong correspondences from indistinct features.
Through using gradient rather than texture characteristics, the results of DSO are acceptable but also drift away when switching the viewpoint.
The thermal results indicate that visible spectral methods cannot be applied directly to a new spectrum.
Compared with DSO and ORB-SLAM2 in monocular setups, the proposed method provides more robust results for most of the sequences since static stereo constraints are employed to ensure that the metric scale is consistent. 
In other words, the proposed method can use thermal data to keep the metric scale for long sequences as shown in Figure~\ref{fig:data_ass_vicon}.
In addition, the results of DSO(RGB) are close to those of the proposed method, implying that RGB information is the main source for motion estimation.

\begin{table}[b]
	
	\centering
	\caption{The evaluation results of Absolute Trajectory Error (ATE) on the multi-spectral dataset[m]. The * means tracking lost in operation and $\times$ means stereo ORB-SLAM2 cannot be initialized. The best results are shown in bold.
	}
	\label{tab_msd}
	\vspace{0.1cm}
	\resizebox{9cm}{!}{
		\begin{tabular}{c|cccccc}
			\toprule
			Method &  MSVO   & ORB-SLAM2(RGB) & ORB-SLAM2(thermal) &ORB-SLAM2(stereo) &  DSO (RGB) &  DSO (thermal)  \\
			\midrule
			sq-01   &{0.0841}   & \textbf{0.0697}    & $\times$ & $\times$ & 0.1079       & 0.2511  \\
			sq-02   &{0.0227} & 0.1183     & $\times$ & $\times$ & \textbf{0.0212}     & 0.2488  \\
			sq-03   &{0.0204}& 0.3521   & $\times$ & $\times$ & \textbf{0.0196}      & 0.4225  \\
			sq-04  & \textbf{0.0600} & {0.1345}     & $\times$ & $\times$ & 0.0618    & 0.4986  \\
			sq-05   & \textbf{0.0282}   & 0.763    & 0.1132 & $\times$ & 0.0316   & {0.2553} \\
			sq-06  & \textbf{0.0418} 	& $\times$  & $\times$ & $\times$ & 0.0490     & 0.1610  \\
			sq-07  & \textbf{0.0848} 	& $\times$  & $\times$ & $\times$ & 0.1060     & 0.2331  \\
			\bottomrule
	\end{tabular}}
	
\end{table}
\subsection{Qualitative evaluation}

\subsubsection{Stereo association}

\begin{figure}[t]
	\centering
	\includegraphics[scale=0.14]{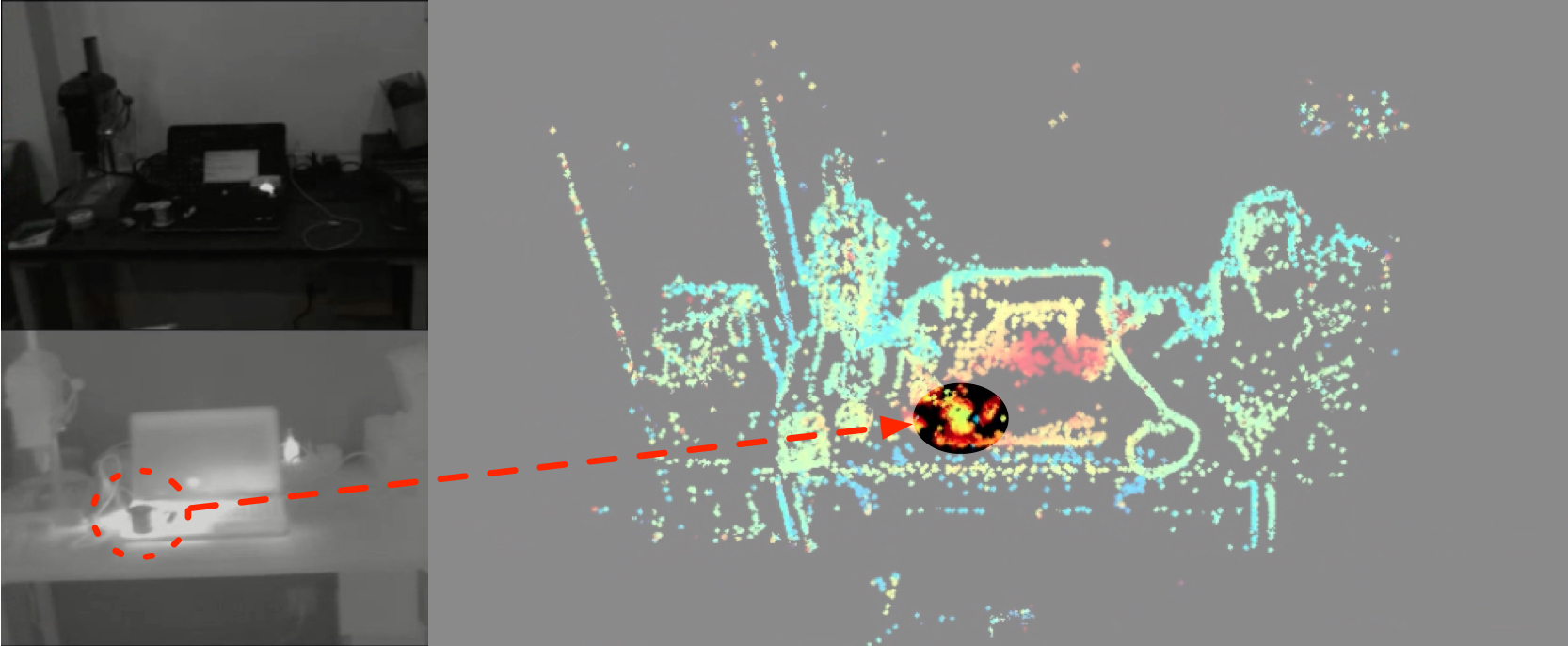}
	\caption{Example of a fused point cloud. The left top picture is the RGB image and the bottom is the thermal image. The right picture shows the point cloud assigned with the thermal information from the left bottom image using point reprojection. The warmer the color, the higher the temperature. The correct correspondences ensure that the cup is around the high temperature region as in the real world.}
	\label{fig:fusion_exp}
\end{figure}

As shown in Figure~\ref{fig:fusion_exp}, an image sequence containing 933 frames with 32 fps was recorded by  moving the multi-spectral device around a table with a laptop on the top of it.
As shown in the left top picture of Figure~\ref{fig:fusion_exp}, the thermal information is fused correctly into the point cloud with correct stereo correspondences.
Note that correct metric scale ensures that the cup is around the high temperature region in the point cloud, which is consistent with the original thermal data.
As a result, each stereo correspondence has two correct intensity references in each corresponding spectrum.
This experiment indicates that the proposed MSVO can fuse different spectral information and recover the metric scale in indoor environments.

\begin{figure}[bpt]
	\centering
	\begin{minipage}[t]{0.2\textwidth}
		\includegraphics[width=\textwidth]  {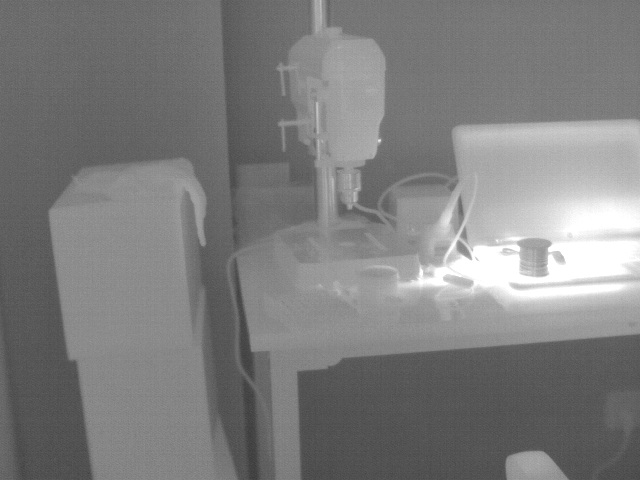}
		\centerline{\footnotesize(a)  Before NUC}
	\end{minipage}
	\begin{minipage}[t]{0.2\textwidth}
		\includegraphics[width=\textwidth]  {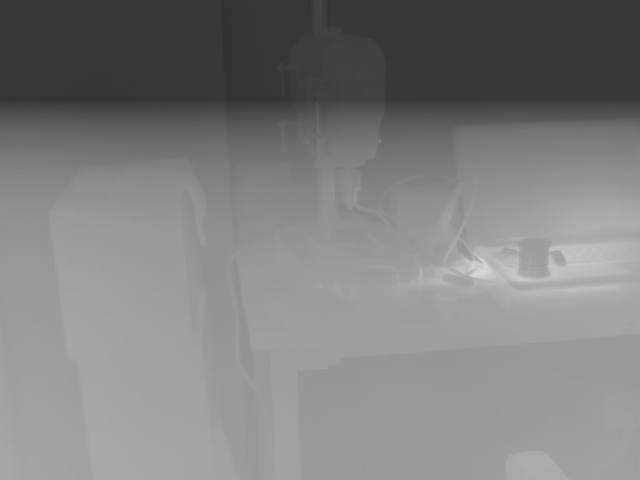}
		\centerline{\footnotesize(b)  During NUC}
	\end{minipage}
	\caption{Normal and corrupted thermal images. If the LWIR camera enters NUC process, the latest data will be corrupted.}
	\label{fig:flag_refresh}
\end{figure}

Inevitably, the LWIR camera needs to refresh for NUC, as shown in Figure~\ref{fig:flag_refresh}. Based on the proposed method, the corrupted data were discarded and the latest pose result was estimated using only the information from the standard camera under the scale constraint.

\subsubsection{Outdoor evaluation}

\begin{figure}[t]
	\centering
	\includegraphics[scale=0.25]{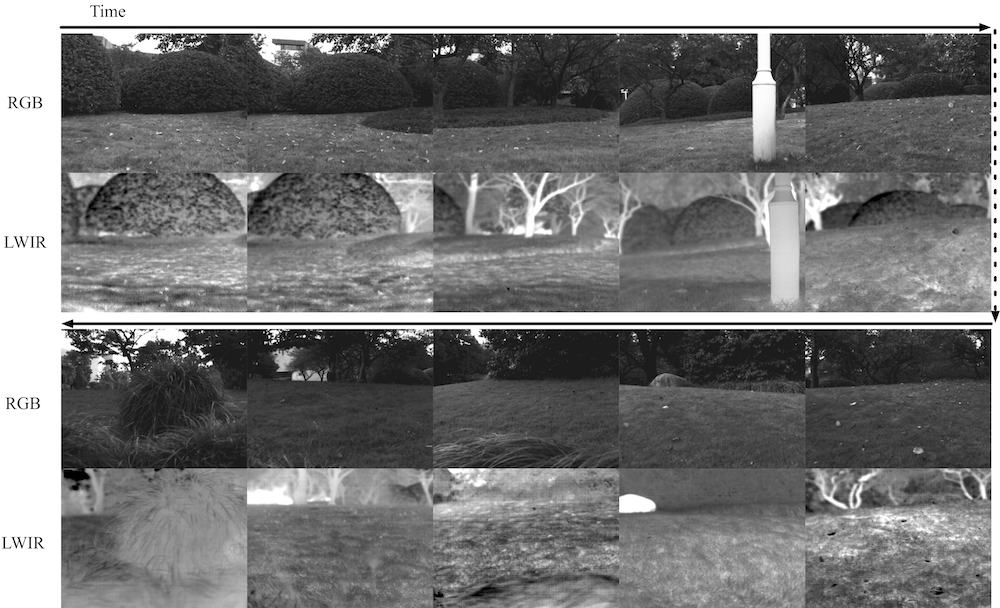}
	\caption{Outdoor multi-spectral sequence.
		The 10 RGB plots show the visible images while
	the 10 LWIR plots show the corresponding thermal images.
	}
	\label{fig:ovofourdoorvideo}
\end{figure}

As shown in Figure~\ref{fig:ovofourdoorvideo}, an image sequence around the boundary of a garden at nightfall was sensed by the multi-spectral device installed on a vehicle, where the edge of the garden can be treated as the groundtruth in the sequence. It contains 9511 frames with 32 fps. In addition to NUC corruption, due to the limited focus range of the LWIR camera, a large amount of high frequency data was lost.
Moreover, the repeating texture also poses additional difficulty for visual odometry in this environment. Since the stereo ORB-SLAM2 cannot be performed on this sequence, GPS (single point positioning system) data is used for performance evaluation. 

\begin{figure}[t]
	\centering
	\includegraphics[scale=0.3]{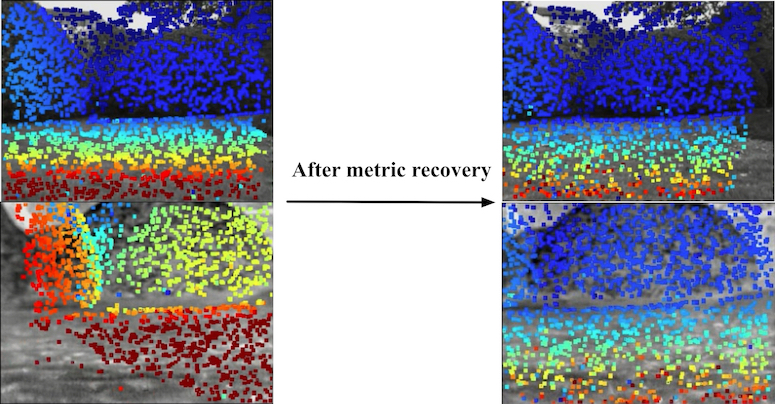}
	\caption{LWIR depth map before and after metric scale recovery.
	}
	\label{fig:data_ass_outdoor}
\end{figure}

\begin{figure}[t]
	\centering
	\includegraphics[scale=0.25]{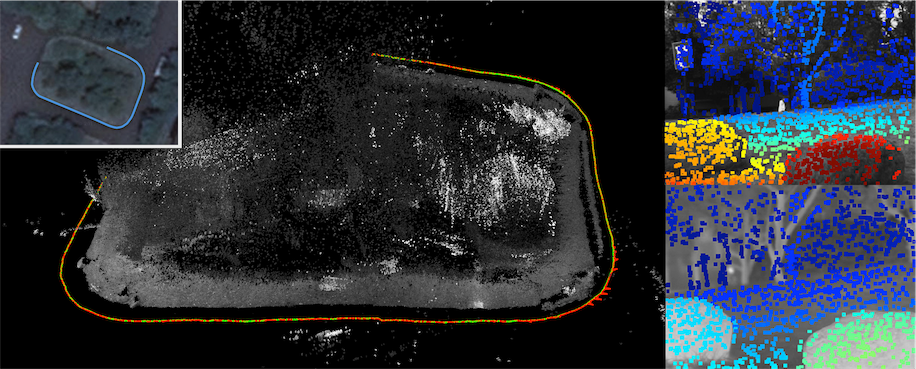}
	\caption{The point cloud and stereo association results.
		The top left image is the satellite image of the campus environment with a blue groundtruth trajectory on it.
		The middle plot gives the overview of the MSVO results. The right plots show depth maps of stereo images.
		The color indicates the distance from the 3D point to the optic center of that camera.
	}
	\label{fig:ms_ov}
\end{figure}

As shown in Figure~\ref{fig:data_ass_outdoor}, the system can recover the metric scale in unstructured environments.
Meanwhile, the right two depth maps in Figure~\ref{fig:ms_ov} created from different cameras indicate that correct stereo correspondences have been  successfully obtained.

\begin{figure}[t]
	\centering
	\includegraphics[scale=0.06]{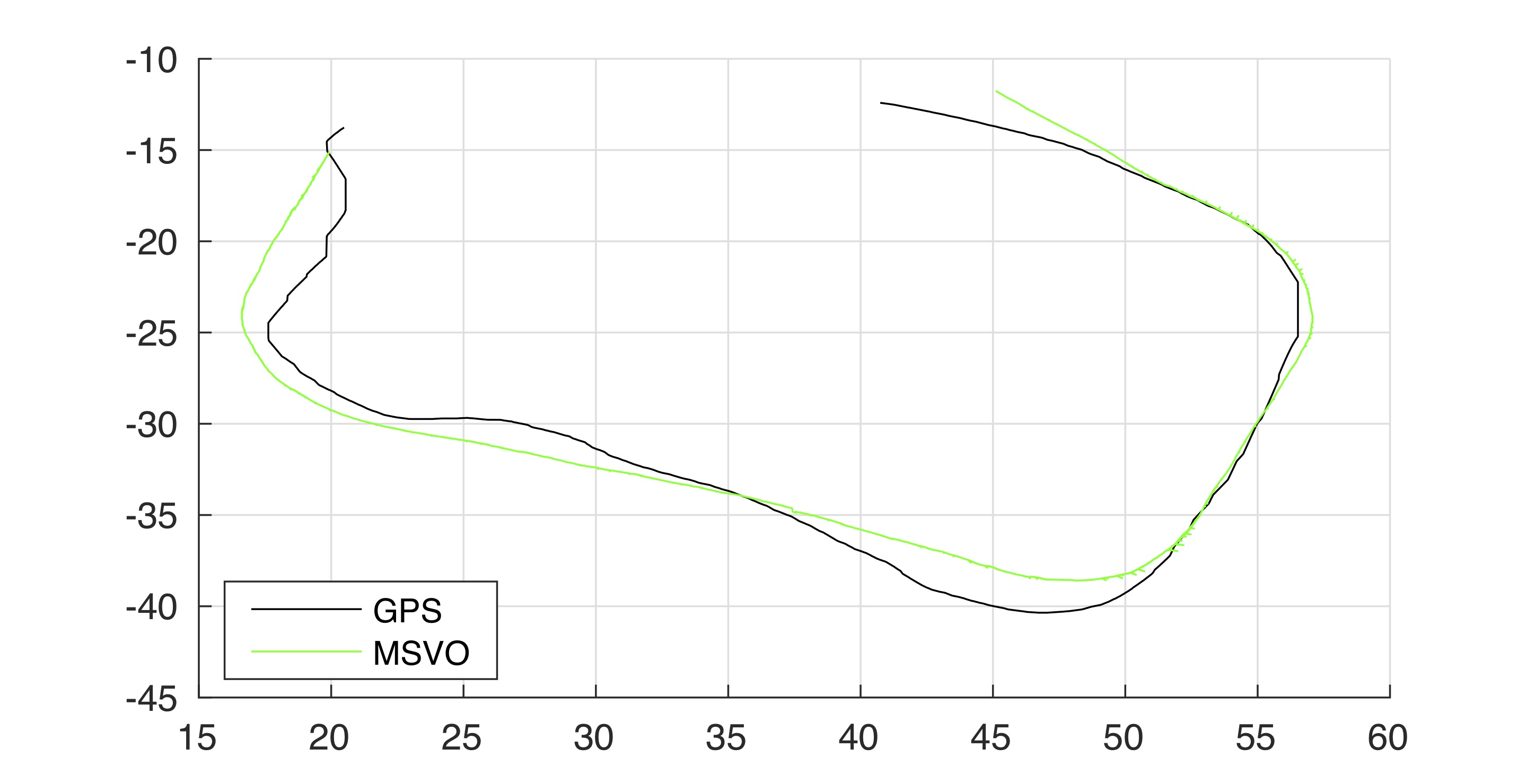}
	\caption{Qualitative result. The estimated trajectory (green) is aligned to GPS results (black) on the outdoor sequence.}
	\label{fig:ms_result}
\end{figure}

To evaluate the algorithm qualitatively, the MSVO result aligned to the GPS output is shown in Figure~\ref{fig:ms_result}. Since the result is closer to the shape of the garden, the proposed method produces more precise estimation than the GPS. 

\section{Conclusion}
\label{conclusion}

In this work, a multi-spectral visual odometry method was proposed that can exploit multi-spectral information without explicit stereo matching. It can reconstruct a semi-dense multi-spectral map even when some points share no similarity in different spectra. The idea of combing the information of static stereo through individual observations was discussed in detail. The proposed method was evaluated on several multi-spectral sequences. Qualitative and quantitative experimental results demonstrate that the proposed framework is capable of fusing multi-view stereo information and providing accurate results without using stereo matching. 

In future work, the special noise on LWIR cameras will be considered to improve the accuracy of the design. In addition, by constructing supplementary map points in the LWIR camera, robustness of the proposed method in environments with highly dynamic illumination conditions can be enhanced.

{

{\small
\bibliographystyle{ieee}
\bibliography{ref}
}
}

\end{document}